\documentclass{article}

\usepackage{microtype}
\usepackage{graphicx}
\usepackage{subfigure}
\usepackage{booktabs} 
\usepackage[utf8]{inputenc} 
\usepackage[T1]{fontenc}    
\usepackage{hyperref}       
\usepackage{url}            
\usepackage{booktabs}       
\usepackage{amsfonts}       
\usepackage{nicefrac}       
\usepackage{microtype}      
\usepackage{subfigure}
\usepackage{float}
\usepackage{wrapfig}
\usepackage{graphicx, caption}
\usepackage{tabularx}

\usepackage{hyperref}

\usepackage[accepted]{icml2020}

\icmltitlerunning{MERL}

\begin{document}

\twocolumn[

\icmltitle{Evolutionary Reinforcement Learning for Sample-Efficient Multiagent Coordination}



\icmlsetsymbol{equal}{*}

\begin{icmlauthorlist}
\icmlauthor{Shauharda Khadka}{equal,org1}
\icmlauthor{Somdeb Majumdar}{equal,org1}
\icmlauthor{Santiago Miret}{org1}
\icmlauthor{Stephen McAleer}{org2}
\icmlauthor{Kagan Tumer}{org3}
\end{icmlauthorlist}

\icmlaffiliation{org1}{Intel Labs}
\icmlaffiliation{org2}{University of California, Irvine}
\icmlaffiliation{org3}{Oregon State University}

\icmlcorrespondingauthor{Shauharda Khadka}{shauharda.khadka@intel.com}
\icmlcorrespondingauthor{Somdeb Majumdar}{somdeb.majumdar@intel.com}

\icmlkeywords{Machine Learning, ICML}

\vskip 0.3in
]



\printAffiliationsAndNotice{\icmlEqualContribution} 

\begin{abstract}
Many cooperative multiagent reinforcement learning environments provide agents with a sparse team-based reward, as well as a dense agent-specific reward that incentivizes learning basic skills. Training policies solely on the team-based reward is often difficult due to its sparsity. Furthermore, relying solely on the agent-specific reward is sub-optimal because it usually does not capture the team coordination objective. A common approach is to use reward shaping to construct a proxy reward by combining the individual rewards. However, this requires manual tuning for each environment. We introduce Multiagent Evolutionary Reinforcement Learning (MERL), a split-level training platform that handles the two objectives separately through two optimization processes. An evolutionary algorithm maximizes the sparse team-based objective through neuroevolution on a population of teams. Concurrently, a gradient-based optimizer trains policies to only maximize the dense agent-specific rewards. The gradient-based policies are periodically added to the evolutionary population as a way of information transfer between the two optimization processes. This enables the evolutionary algorithm to use skills learned via the agent-specific rewards toward optimizing the global objective. Results demonstrate that MERL significantly outperforms state-of-the-art methods, such as MADDPG, on a number of difficult coordination benchmarks. 
\end{abstract}

\section{Introduction}

Cooperative multiagent reinforcement learning (MARL) studies how multiple agents can learn to coordinate as a team toward maximizing a global objective. Cooperative MARL has been applied to many real world applications such as air traffic control \citep{tumer2007distributed}, multi-robot coordination \citep{sheng2006distributed,yliniemi2014multirobot}, communication and language \citep{lazaridou2016multi, mordatch2018emergence}, and autonomous driving \citep{shalev2016safe}. 

Many such environments endow agents with a team reward that reflects the team's coordination objective, as well as an agent-specific local reward that rewards basic skills. For instance, in soccer, dense local rewards could capture agent-specific skills such as passing, dribbling and running. The agents must then coordinate when and where to use these skills in order to optimize the team objective, which is winning the game. Usually, the agent-specific reward is dense and easy to learn from, while the team reward is sparse and requires the cooperation of all or most agents. 

Having each agent directly optimize the team reward and ignore the agent-specific reward usually fails or leads to sample-inefficiency for complex tasks due to the sparsity of the team reward. Conversely, having each agent directly optimize the agent-specific reward also fails because it does not capture the team's objective, even with state-of-the-art methods such as MADDPG \citep{lowe2017multi}.

One approach to this problem is to use reward shaping, where extensive domain knowledge about the task is used to create a proxy reward function \citep{rahmattalabi2016d++}. Constructing this proxy reward function is challenging in complex environments, and is usually domain-dependent. In addition to requiring domain knowledge and manual tuning, this approach also poses risks of changing the underlying problem itself \citep{ng1999policy}. Common approaches to creating proxy rewards via linear combinations of the two objectives also fail to solve or generalize to complex coordination tasks \citep{ devlin2011multi,williamson2009reward}.   

In this paper, we introduce Multiagent Evolutionary Reinforcement Learning (MERL), a state-of-the-art algorithm for cooperative MARL that does not require reward shaping. MERL is a split-level training platform that combines gradient-based and gradient-free optimization. The gradient-free optimizer is an evolutionary algorithm that maximizes the team objective through neuroevolution. The gradient-based optimizer is a policy gradient algorithm that maximizes each agent's dense, local rewards. These gradient-based policies are periodically copied into the evolutionary population, while the two processes operate concurrently and share information through a shared replay buffer.

A key strength of MERL is that it is a general method which does not require domain-specific reward shaping. This is because MERL optimizes the team objective directly while simultaneously leveraging agent-specific rewards to learn basic skills. We test MERL in a number of multiagent coordination benchmarks. Results demonstrate that MERL significantly outperforms state-of-the-art methods such as MADDPG, while using the same observations and reward functions. We also demonstrate that MERL scales gracefully to increasing complexity of coordination objectives where MADDPG and its variants fail to learn entirely.

\section{Background and Related Work}
\label{sec:back}

\textbf{Markov Games:} A standard reinforcement learning (RL) setting is formalized as a Markov Decision Process (MDP) and consists of an agent interacting with an environment over a finite number of discrete time steps. This formulation can be extended to multiagent systems in the form of partially observable Markov games \citep{littman1994markov}. An $N$-agent Markov game is defined by a global state of the world, $\mathcal{S}$, and a set of $N$ observations $\{\mathcal{O}_i\}$ and $N$ actions $\{\mathcal{A}_i\}$ corresponding to the $N$ agents. At each time step $t$, each agent observes its corresponding observation $O_i^t$ and maps it to an action $A_i^t$ using its policy $\pi_i$. 

Each agent receives a scalar reward $r_i^t$ based on the global state $\mathcal{S}_t$ and joint action of the team. 
The world then transitions to the next state $\mathcal{S}_{t+1}$ which produces a new set of observations $\{\mathcal{O}_i\}$. The process continues until a terminal state is reached. $R_i = \sum_{t=0}^{T} \gamma^tr_i^t$ is the total return for agent $i$ with discount factor $\gamma \in (0, 1]$. Each agent aims to maximize its expected return. 

\textbf{TD3:} Policy gradient (PG) methods frame the goal of maximizing the expected return as the minimization of a loss function. 
A widely used PG method for continuous, high-dimensional action spaces is DDPG \citep{lillicrap2015}. Recently, \citep{fujimoto2018addressing} extended DDPG to Twin Delayed DDPG (TD3), addressing its well-known overestimation problem. TD3 is the state-of-the-art, off-policy algorithm for model-free DRL in continuous action spaces.

TD3 uses an actor-critic architecture \citep{sutton1998} maintaining a deterministic policy (actor) $\pi: \mathcal{S} \rightarrow \mathcal{A}$, and two distinct critics $\mathcal{Q}: \mathcal{S} \times \mathcal{A} \rightarrow \mathbb{R}_i$. Each critic independently approximates the actor's action-value function $\mathcal{Q}^\pi$. A separate copy of the actor and critics are kept as target networks for stability and are updated periodically. A noisy version of the actor is used to explore the environment during training. The actor is trained using a noisy version of the sampled policy gradient computed by backpropagation through the combined actor-critic networks. This mitigates overfitting of the deterministic policy by smoothing the policy gradient updates.  

\textbf{Evolutionary Reinforcement Learning (ERL)} is a hybrid algorithm that combines Evolutionary Algorithms (EAs) \citep{floreano2008, luders2017,fogel2006,spears1993}, with policy gradient methods \citep{khadka2018evolution}. Instead of discarding the data generated during a standard EA rollout, ERL stores this data in a central replay buffer shared with the policy gradient's own rollouts - thereby increasing the diversity of the data available for the policy gradient learners. Since the EA directly optimizes for episode-wide return, it biases exploration towards states with higher long-term returns. The policy gradient algorithm which learns using this state distribution inherits this implicit bias towards long-term optimization. Concurrently, the actor trained by the policy gradient algorithm is inserted into the evolutionary population allowing the EA to benefit from the fast gradient-based learning.    
 
\textbf{Related Work}: \cite{lowe2017multi} introduced MADDPG which tackled the inherent non-stationarity of a multiagent learning environment by leveraging a critic which had full access to the joint state and action during training. \cite{foerster2018counterfactual} utilized a similar setup with a centralized critic across agents to tackle StarCraft micromanagement tasks. An algorithm that could explicitly model other agents' learning was investigated in \cite{foerster2018learning}. However, all these approaches rely on a dense agent reward that properly captures the team objective. Methods to solve for these types of agent-specific reward functions were investigated in \cite{li2012optimal} but were limited to tasks with strong simulators where tree-based planning could be used. 

A closely related work to MERL is \citep{liu2019emergent} where Population-Based Training (PBT) \citep{jaderberg2017} is used to optimize the relative importance between a collection of dense, shaped rewards alongside their discount rates automatically during training. This can be interpreted as a singular central reward function constructed by scalarizing a collection of reward signals where the scalarization coefficients and discount rates are adaptively learned during training. PBT-MARL a powerful method that can leverage complex mixtures of its reward signals throughout training. However, it still relies on finding ideal mixing function between rewards signals to drive learning. In contrast, MERL optimizes its reward functions independently, relying instead on information transfer across them to drive learning. This is facilitated through shared replay buffers and policy migration directly. This form of information transfer through a shared replay buffer has been explored extensively in recent literature \citep{colas2018, khadka2019cerl}.

\section{Motivating Example}
\setlength\intextsep{0pt}

\begin{figure}[ht]

\subfigure[Rover domain]{
\includegraphics[width=0.48\columnwidth]{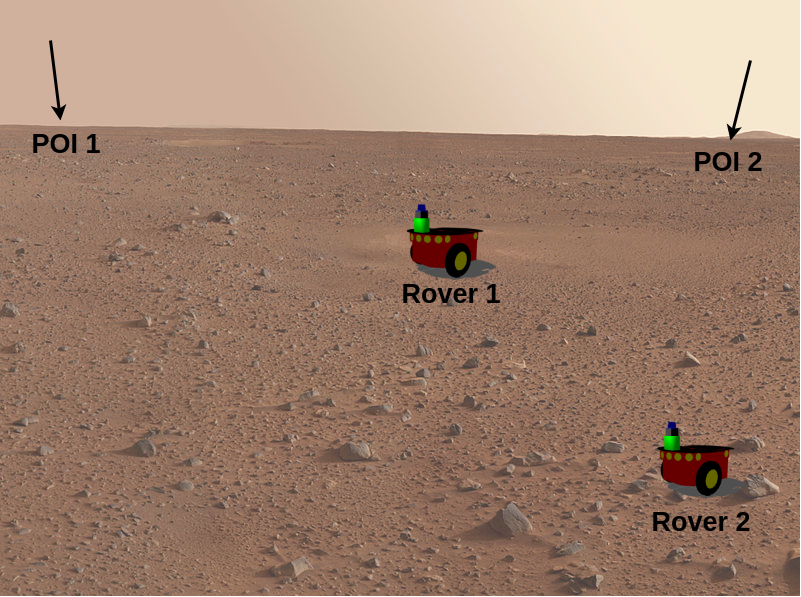}}\hspace*{\fill}
\subfigure[MERL vs TD3 vs EA]{
\includegraphics[width=0.5\columnwidth]{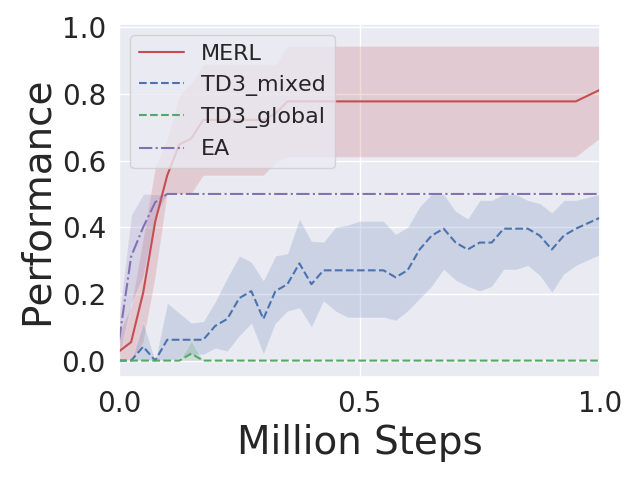}}\hspace*{\fill}
\caption{(a) Rover domain with a clear misalignment between agent and team reward functions (b) Comparative performance of MERL compared against TD3-mixed, TD3-global and EA.}
\label{fig:motivate}
\vspace{1em}
\end{figure}

Consider the rover domain \citep{agogino2004unifying}, a classic multiagent task where a team of rovers coordinate to explore a region. The team objective is to observe all POIs (Points of Interest). Each robot also receives an agent-specific reward defined as negative distance to the closest POI. In Figure \ref{fig:motivate}(a), a team of two rovers $R_1$ and $R_2$ seek to explore and observe POIs $P_1$ and $P_2$. $R_1$ is closer to $P_2$ and has enough fuel to reach either of the POIs whereas $R_2$ can only reach $P_2$. There is no communication between the rovers.

If $R_1$ optimizes only locally by pursuing the closer POI $P_2$, then the team objective is not achieved since $R_2$ can only reach $P_2$. The globally optimal solution for $R_1$ is to spend more fuel and pursue $P_1$, which is misaligned with its locally optimal solution. Figure \ref{fig:motivate}(b) shows the comparative performance of four algorithms - namely TD3-global, TD3-mixed, EA and MERL on this coordination task.

\textbf{TD3-global} optimizes only the team reward and \textbf{TD3-mixed} optimizes a linear mixture of the individual and team rewards. Since the team reward is sparse, TD3-global fails to learn anything meaningful. In contrast, TD3-mixed, due to the dense agent-specific reward component, successfully learns to perceive and navigate. However, since the mixed reward does not capture the true team objective, it converges to the greedy local policy of $R_1$ pursuing $P_2$.

\textbf{EA} relies on randomly stumbling onto a solution - e.g., a trajectory that takes the rover to a POI. Since EA directly optimizes the team objective, it has a strong preference for the globally optimal solution. However, the probability of the rovers randomly stumbling onto the globally optimal solution is extremely low. The greedy solution is significantly more likely to be found - and is what EA converges to.

\textbf{MERL} combines the core strengths of TD3 and EA. The TD3 component in MERL exploits the dense agent-specific reward to learn perception and navigation skills while being agnostic to the global objective. The task is then reduced to its coordination component - picking the right POI to go to. This is effectively tackled by the EA module within MERL. This ability to independently leverage reward functions across multiple levels even when they are misaligned is the core strength of MERL.

\section{Multiagent Evolutionary Reinforcement Learning}

MERL leverages both agent-specific and team objectives through a hybrid algorithm that combines gradient-free and gradient-based optimization. The gradient-free optimizer is an evolutionary algorithm that maximizes the team objective through neuroevolution. The gradient-based optimizer trains policies to maximize agent-specific rewards. These gradient-based policies are periodically added to the evolutionary population and participate in evolution. This enables the evolutionary algorithm to use agent-specific skills learned by training on the agent-specific rewards toward optimizing the team objective without needing to resort to reward shaping.

\begin{figure}[h]
\begin{center}
\centerline{\includegraphics[width=0.7\columnwidth]{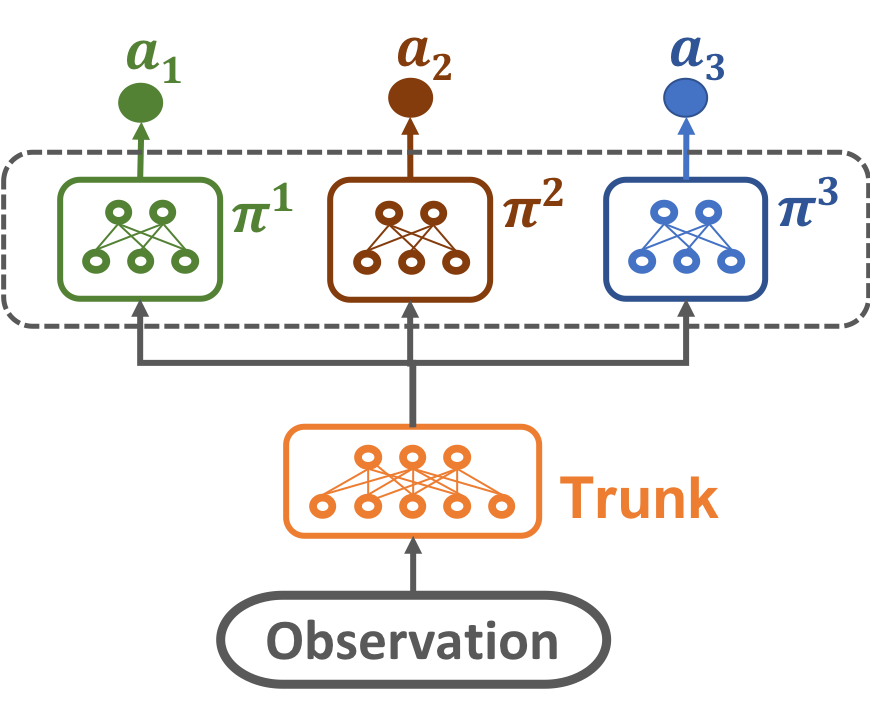}}
\caption{Multi-headed team policy}
\label{fig:multihead}
\end{center}
\end{figure}

\textbf{Policy Topology:} We represent our multiagent (\textit{team}) policies using a multi-headed neural network $\pi$ as illustrated in Figure \ref{fig:multihead}. The head $\pi^k$ represents the $k$-th agent in the team. Given an incoming observation for agent $k$, only the output of $\pi^k$ is considered as agent $k$'s response. In essence, all agents act independently based on their own observations while sharing weights (and by extension, the features) in the lower layers (\textit{trunk}). This is commonly used to improve learning speed \citep{silver2017mastering}. Further, each agent $k$ also has its own replay buffer $\mathcal(R^k)$ which stores its \textit{experience} defined by the tuple \textit{(state, action, next state, local reward)} for each interaction with the environment (\textit{rollout}) involving that agent.

\begin{figure}[h]
\vskip 0.1in
\begin{center}
\centerline{\includegraphics[width=0.99\columnwidth]{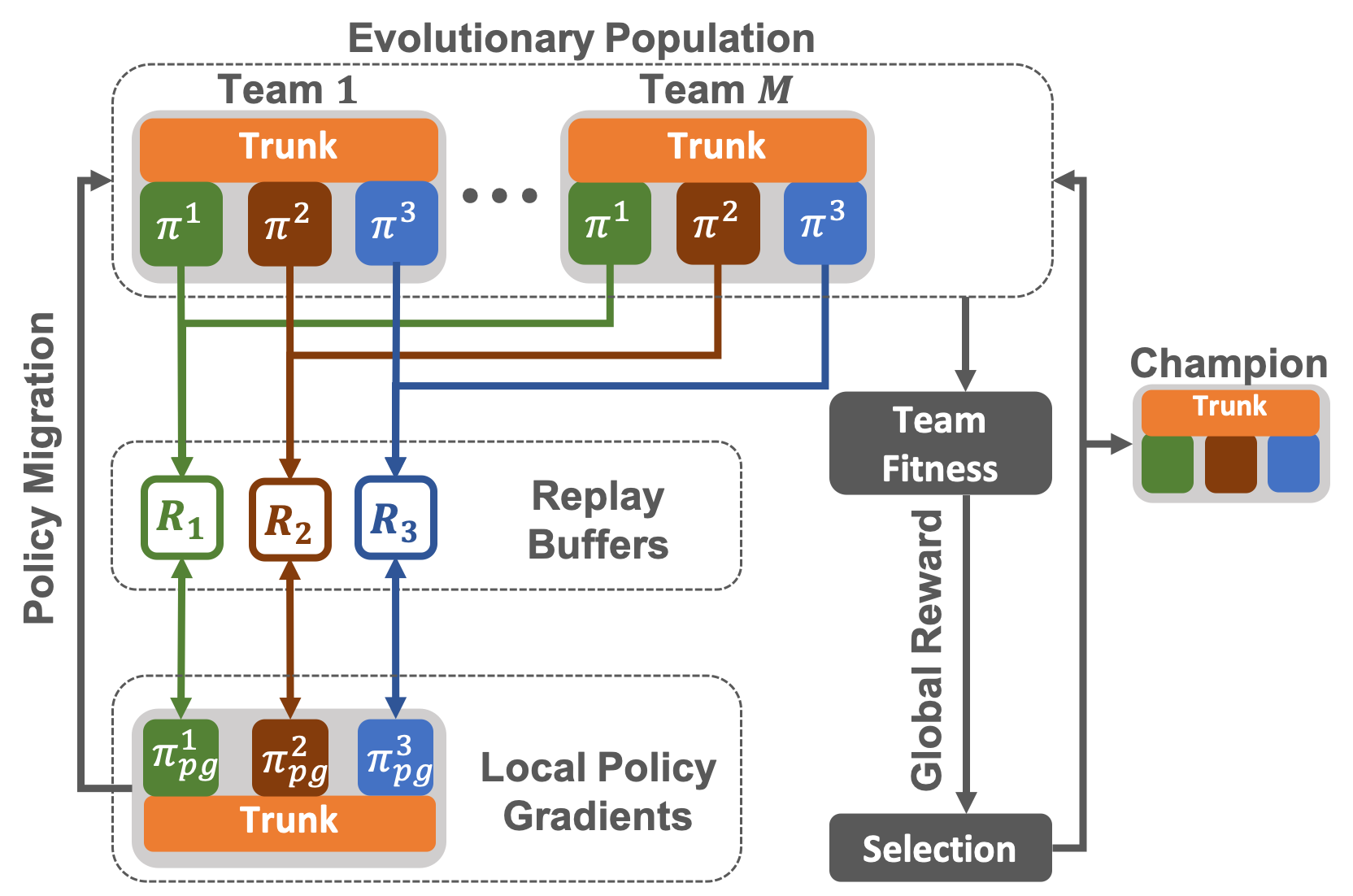}}
\caption{High level schematic of MERL highlighting the integration of local and global reward functions}
\label{fig:merl_process}
\end{center}
\vskip -0.1in
\end{figure}

\textbf{Team Reward Optimization:} Figure \ref{fig:merl_process} illustrates the MERL algorithm. A population of $M$ multi-headed teams, each with the same topology, is initialized with random weights. The replay buffer $\mathcal{R}^k$ is shared by the $k$-th agent across all teams. The population is then \textit{evaluated} for each rollout. The team reward for each team is disbursed at the end of the episode and is considered as its \textbf{fitness score}. A \textbf{selection} operator selects a portion of the population for survival with probability proportionate to their fitness scores. The weights of the teams in the population are probabilistically \textit{perturbed} through mutation and crossover operators to create the next \textit{generation} of teams. A portion of the teams with the highest relative fitness is preserved as elites. At any given time, the team with the highest fitness, or the \textit{champion}, represents the best solution for the task.

\textbf{Policy Gradient:} The procedure described so far resembles a standard EA except that each agent $k$ stores each of its experiences in its associated replay buffer $\mathcal(R^k)$ instead of just discarding it. However, unlike EA, which only learns based on the low-fidelity global reward, MERL also learns from the experiences within episodes of a rollout using policy gradients. To enable this kind of "local learning", MERL initializes one multi-headed policy network $\mathcal{\pi}_{pg}$ and one critic $\mathcal{Q}$. A noisy version of $\mathcal{\pi}_{pg}$ is then used to conduct its own set of rollouts in the environment, storing each agent $k$'s experiences in its corresponding buffer $\mathcal(R^k)$ similar to the evolutionary rollouts.

\textbf{Agent-Specific Reward Optimization:} Crucially, each agent's replay buffer is kept separate from that of every other agent to ensure diversity amongst the agents. The shared critic samples a random mini-batch uniformly from each replay buffer and uses it to update its parameters using gradient descent. Each agent $\pi^k_{pg}$ then draws a mini-batch of experiences from its corresponding buffer $\mathcal(R^k)$ and uses it to sample a policy gradient from the shared critic. Unlike the teams in the evolutionary population which directly seek to optimize the team reward, $\pi_{pg}$ seeks to maximize the agent-specific local reward while exploiting the experiences collected via evolution.

\textbf{Skill Migration:} Periodically, the $\mathcal{\pi}_{pg}$ network is copied into the evolving population of teams and propagates its features by participating in evolution. This is the core mechanism that combines policies learned via agent-specific and team rewards. Regardless of whether the two rewards are aligned, evolution ensures that only the performant derivatives of the migrated network are retained. This mechanism guarantees protection against destructive interference commonly seen when a direct scalarization between two reward functions is attempted. Further, the level of information exchange is automatically adjusted during the process of learning, in contrast to being manually tuned by an expert.

\vspace{2em}
\begin{algorithm}[]
    \caption{MERL}
    \label{alg:merl} 
\begin{algorithmic}[1]
\STATE Initialize a population of $M$ multi-head teams $pop_{\pi}$ with $k$ agents each and initialize their weights $\theta^\pi$ 
\STATE Initialize a shared critic $\mathcal{Q}$ with weights $\theta^\mathcal{Q}$  
\STATE Initialize an ensemble of $N$ empty cyclic replay buffers ${\mathcal{R}^k}$, one for each agent
\STATE Define a white Gaussian noise generator $\mathcal{W}_g$ random number generator $r()$ $\in$ $[0,1)$ 
\FOR{generation = 1, $\infty$}
    
    \FOR{team $\pi$ $\in$ $pop_{\pi}$}
        \STATE $g$, $\mathcal{R}$ = Rollout ($\pi$, $\mathcal{R}$, noise=None, $\xi$)
        \STATE $\_,$ $\mathcal{R}$ = Rollout ($\pi$, $\mathcal{R}$, noise=$\mathcal{W}_g$, $\xi=1$)
        \STATE Assign $g$ as $\pi$'s fitness

    \ENDFOR

    \STATE Rank the population $pop_{\pi}$ based on fitness scores 
    \STATE Select the first $e$ teams $\pi$ $\in$ $pop_{\pi}$ as elites
    \STATE Select the remaining $(M-e)$ teams $\pi$ from $pop_{\pi}$, to form Set $S$ using tournament selection 
    \WHILE{$|S|$ $<$ $(M-e)$}
         \STATE Single-point crossover between a randomly sampled $\pi \in e$ and $\pi \in S$ and append to $S$
    \ENDWHILE

    \FOR{Agent $k$=$1$,$N$}
    
        \STATE Randomly sample a minibatch of $T$ transitions $(o_i, a_i, l_i, o_{i+1})$ from $R^k$ 
        
                    
        \STATE Compute $y_i$ = $l_i$ + $\gamma \displaystyle \min_{j=1,2} \mathcal{Q}_j'(o_{i+1}, a^\sim|\theta^{\mathcal{Q}_j'})$
        
        \STATE where $a^\sim$ = ${\pi^\prime_{pg}}(k, o_{i+1}|\theta^{\pi_{pg}'})$ [action sampled from the $k^{th}$ head of $\pi_{pg}'$]  $+\epsilon$ 

        \STATE Update $\mathcal{Q}$ by minimizing the loss: $L = \frac{1}{T} \sum_{i} (y_i - \mathcal{Q}(o_i, a_i|\theta^{\mathcal{Q}})^2$ 
        
        \STATE Update $\pi_{pg}^k$  using the sampled policy gradient 
      \begin{center}
        \STATE $\nabla_{\theta^\pi_{pg}}J\sim  \frac{1}{T} \sum \nabla_a\mathcal{Q}(o,a|\theta^\mathcal{Q})|_{o=o_i, a=a_i} \nabla_{\theta^\pi_{pg}} \pi_{pg}^k(s|\theta^\pi_{pg})|_{o=o_i}$
        \end{center}
        
        \STATE Soft update target networks: 
        \STATE $\theta^{\pi'} \Leftarrow \tau \theta^\pi + (1 - \tau)\theta^{\pi'}$ 
        \STATE $\theta^{\mathcal{Q}'} \Leftarrow \tau \theta^\mathcal{Q} + (1 - \tau)\theta^{\mathcal{Q}'}$

    \ENDFOR
        \STATE Migrate the policy gradient team $pop_j:$ for weakest $\pi \in pop_{\pi}^j: \theta^\pi \Leftarrow \theta^{\pi_{pg}}$
            
\ENDFOR
\end{algorithmic}
\end{algorithm}
\vspace{2em}

\begin{figure*}[]
\centering
\subfigure[Predator-Prey]{
\includegraphics[width=0.44\columnwidth]{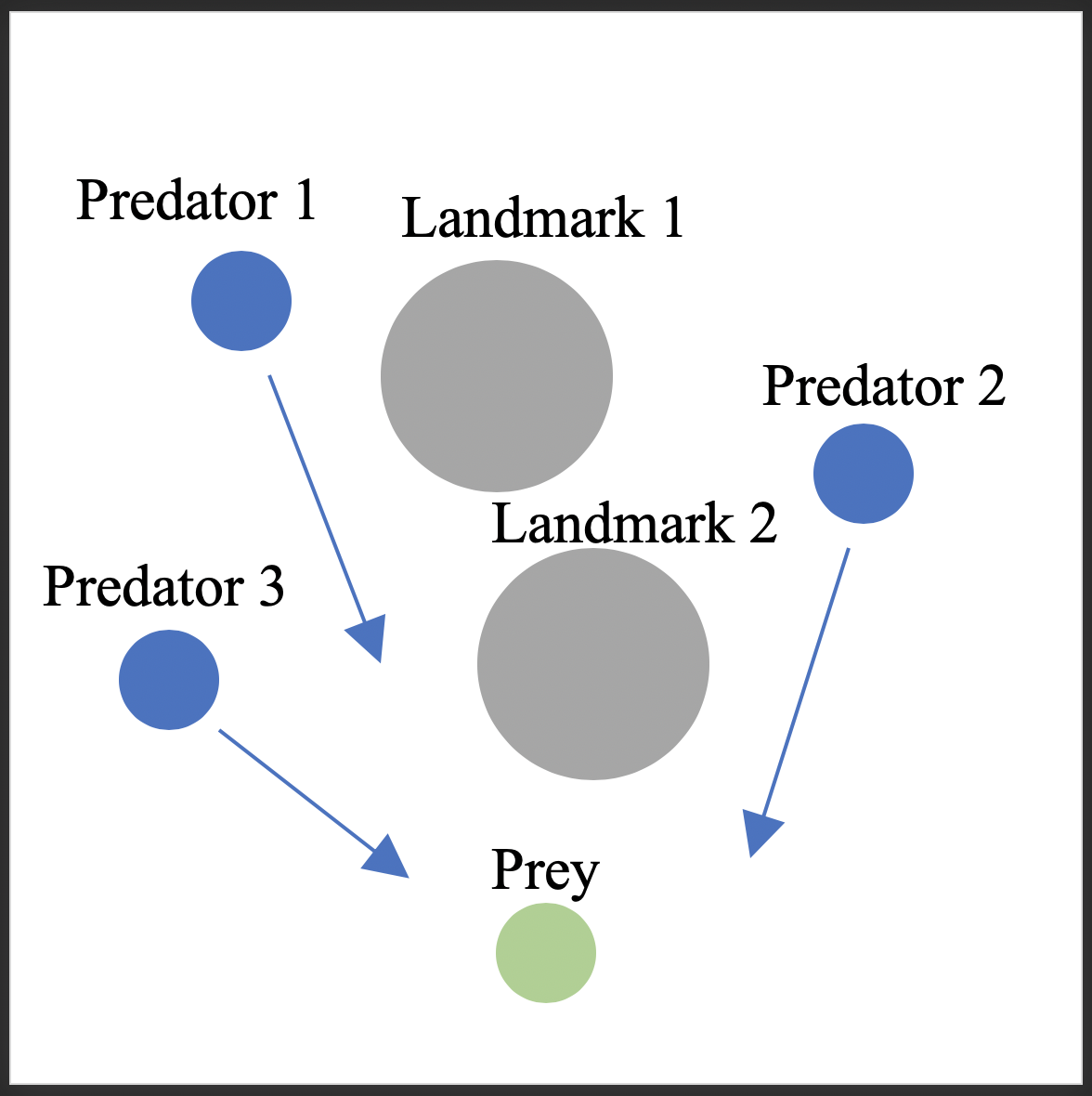}}
\subfigure[Physical Deception]{
\includegraphics[width=0.44\columnwidth]{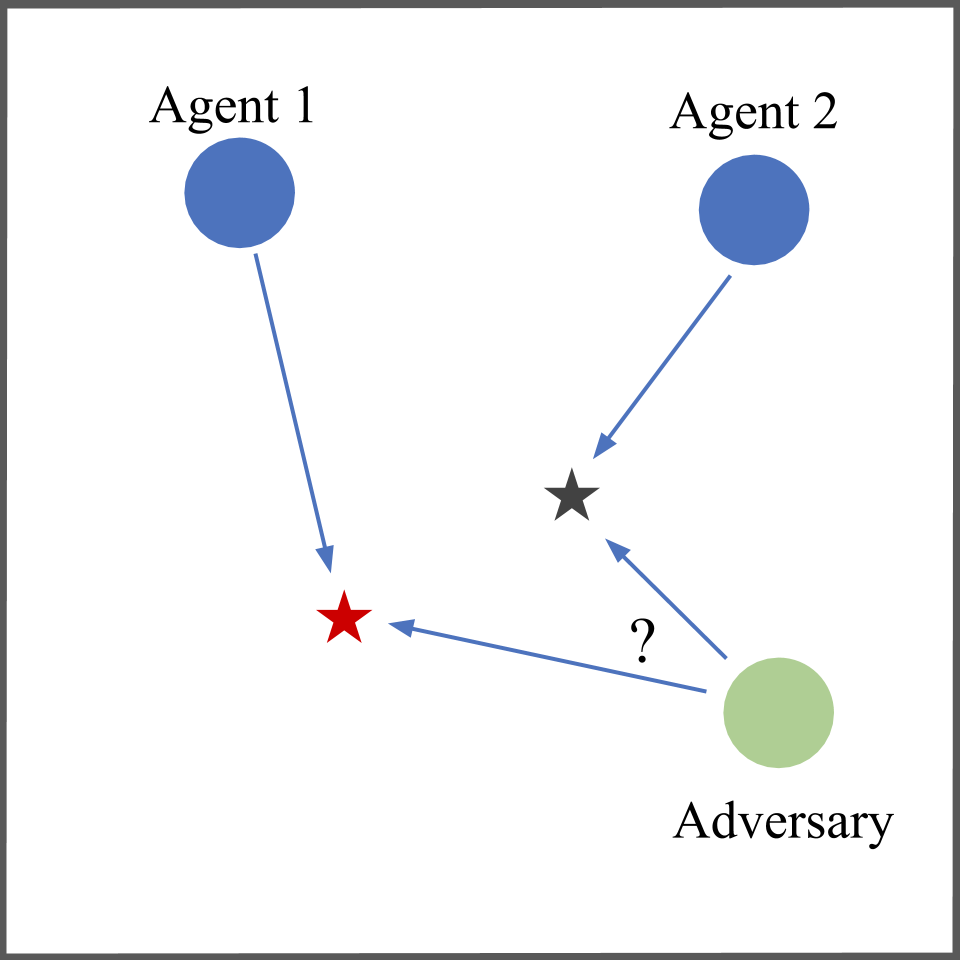}}
\subfigure[Keep-Away]{
\includegraphics[width=0.44\columnwidth]{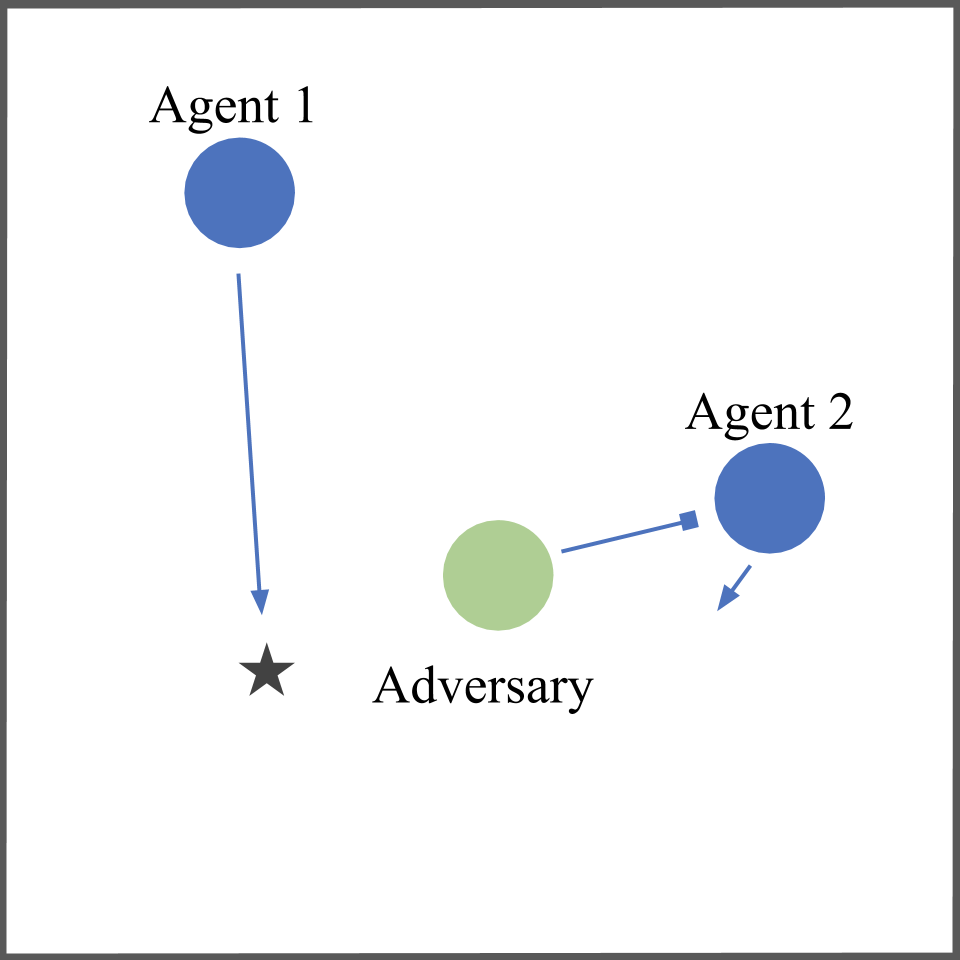}}
\subfigure[Rover Domain]{
\includegraphics[width=0.44\columnwidth]{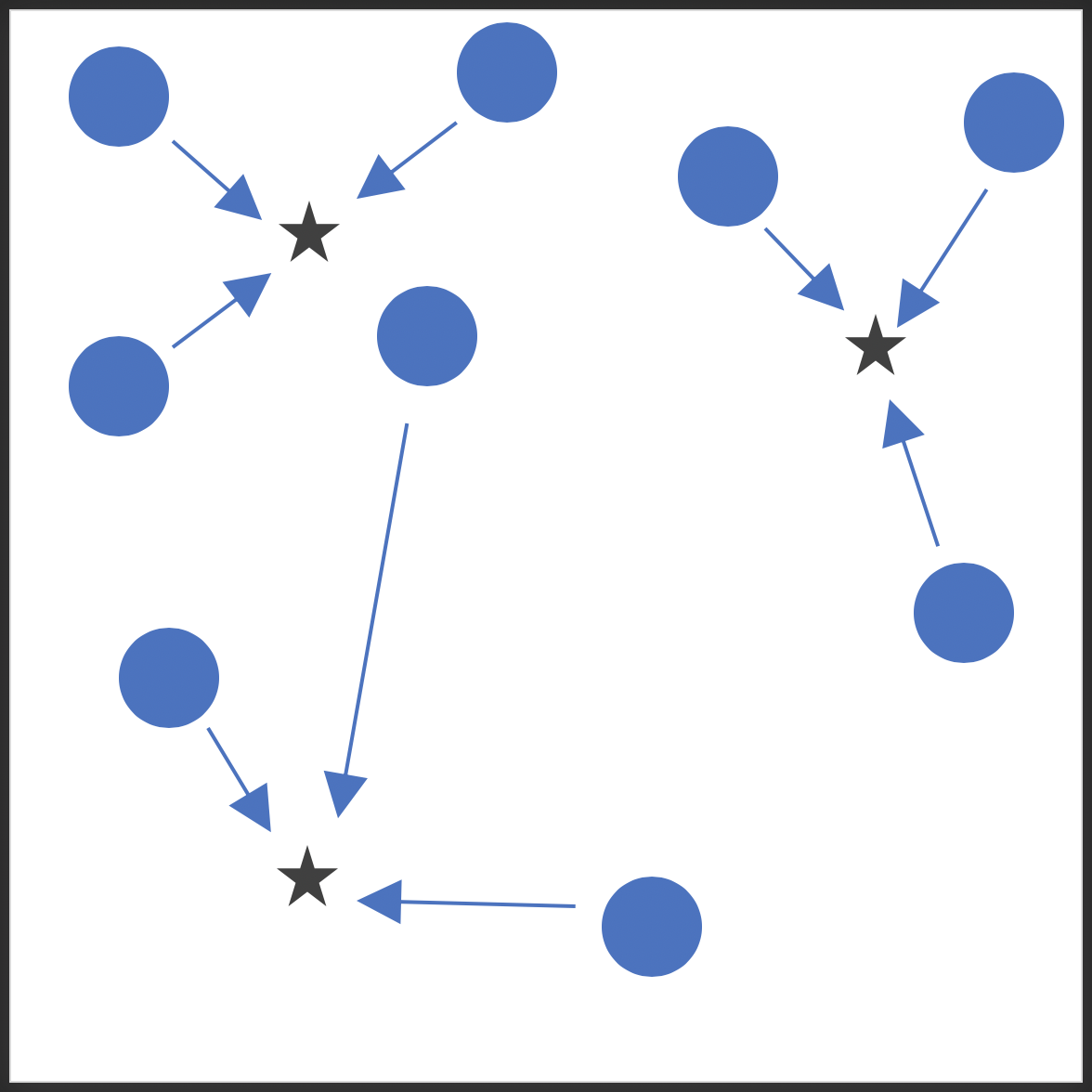}}
\caption{Environments tested  \citep{lowe2017multi,rahmattalabi2016d++}}
\label{fig:envs}
\end{figure*}

Algorithm \ref{alg:merl} provides a detailed pseudo-code of the MERL algorithm. The choice of hyperparameters is explained in the Appendix. Additionally, our source code \footnote{\href{https://tinyurl.com/y6erclts}{https://tinyurl.com/y6erclts}} is available online.

\section{Experiments}

We adopt environments from \citep{lowe2017multi} and \citep{rahmattalabi2016d++} to perform our experiments. Each environment consists of multiple agents and landmarks in a two-dimensional world. Agents take continuous control actions to move about the world. Figure \ref{fig:envs} illustrates the four environments which are described in more detail below.

\textbf{Predator-Prey}:
$N$ slower cooperating agents (predators) must chase the faster adversary (prey) around an environment with $L$ large landmarks in randomly-generated locations. The predators get a reward when they catch (touch) the prey while the prey is penalized. The team reward for the predators is the cumulative number of prey-touches in an episode. Each predator can also compute the average distance to the prey and use it as its agent-specific reward. All agents observe the relative positions and velocities of the other agents, as well as the positions of the landmarks. The prey can accelerate $33\%$ faster than the predator and has a higher top speed. We tests two versions termed simple and hard predator-prey where the prey is $30\%$ and $100\%$ faster, respectively. Additionally, the prey itself learns dynamically during training. We use DDPG \citep{lillicrap2015} as a learning algorithm for training the prey policy. All of our candidate algorithms are tested on their ability to train the team of predators in catching this prey. 

\textbf{Physical Deception}:
 $N$ agents cooperate to reach a single target Point of Interest (POI) among $N$ POIs. They are rewarded based on the closest distance of any agent to the target. A lone adversary also desires to reach the target POI. However, the adversary does not know which of the POIs is the correct one. Thus the cooperating agents must learn to spread out and cover all POIs so as to deceive the adversary as they are penalized based on the adversary's distance to the target. The team reward for the agents is then the cumulative reward in an episode. We use DDPG \citep{lillicrap2015} to train the adversary policy.

\textbf{Keep-Away}: In this scenario, a team of $N$ cooperating agents must reach a target POI out of $L$ total POIs. Each agent is rewarded based on its distance to the target. We construct the team reward as simply the sum of the agent-specific rewards in an episode. An adversary also has to occupy the target while keeping the cooperating agents from reaching the target by pushing them away. To incentivize this behavior, the adversary is rewarded based on its distance to the target POI and penalized based on the distance of the target from the nearest cooperating agent. Additionally, it does not know which of the POIs is the target and must infer this from the movement of the agents.  DDPG \citep{lillicrap2015} is used to train the adversary policy. 

\textbf{Rover Domain}:
This environment is adapted from \citep{rahmattalabi2016d++}. Here, $N$ agents must cooperate to reach a set of $K$ POIs. Multiple agents need to simultaneously go to the same POI in order to observe it. The number of agents required to observe a POI is termed the coupling requirement. Agents do not know and must infer the coupling factor from the rewards obtained. If a team with fewer agents than this number go to a POI, no reward is observed. The team's reward is the percentage of POIs observed at the end of an episode. 

 Each agent can also locally compute its distance to its closest POI and use it as its agent-specific reward. Its observation comprises two channels to detect POIs and rovers, respectively. Each channel receives intensity information over $10^\circ$ resolution spanning the $360^\circ$ around the agent's position loosely based on the characteristic of a Pioneer robot \citep{thrun2000real}. This is similar to a LIDAR. Since it returns the closest reflector, occlusions make the problem partially-observable. A coupling factor of $1$ is similar to the cooperative navigation task in \cite{lowe2017multi}. We test coupling factors from $1$ to $7$ to capture extremely complex coordination objectives.

\textbf{Compared Baselines:} We compare the performance of MERL with a standard neuroevolutionary algorithm (EA) \citep{fogel2006}, MADDPG \citep{lowe2017multi} and MATD3, a variant of MADDPG that integrates the improvements described within TD3 \citep{fujimoto2018addressing} over DDPG. Internally, MERL uses EA and TD3 as its team-reward and agent-specific reward optimizer, respectively. MADDPG was chosen as it is the state-of-the-art multiagent RL algorithm. We implemented MATD3 to ensure that the differences between MADDPG and MERL do not originate from having the more stable TD3 over DDPG. 

Further, we implement MADDPG and MATD3 using either only global team reward or mixed rewards where the local, team-specific rewards were added to the global reward. The local reward function was simply defined as the negative of the distance to the closest POI. In this setting, we sweep over different scaling factors to weigh the two rewards - an approach commonly used to shape rewards.  These experimental variations allow us to evaluate the efficacy of the differentiating features of MERL as opposed to improvements that might come from other ways of combining reward functions.

\textbf{Methodology for Reported Metrics:} 
For MATD3 and MADDPG, the team network was periodically tested on $10$ task instances without any exploratory noise. The average score was logged as its performance. For MERL and EA, the team with the highest fitness was chosen as the champion for each generation. The champion was then tested on $10$ task instances, and the average score was logged. This protocol shielded the reported metrics from any bias of the population size. We conduct $5$ statistically independent runs with random seeds from $\{2019, 2023\}$ and report the average with error bars showing a $95\%$ confidence interval. All scores reported are compared against the number of environment steps (frames). A step is defined as the multiagent team taking a joint action and receiving a feedback from the environment. To make the comparisons fair across single-team and population-based algorithms, all steps taken by all teams in the population are counted cumulatively.
\section{Results}

\begin{figure}[h]
\subfigure[]{
\includegraphics[width=0.48\columnwidth]{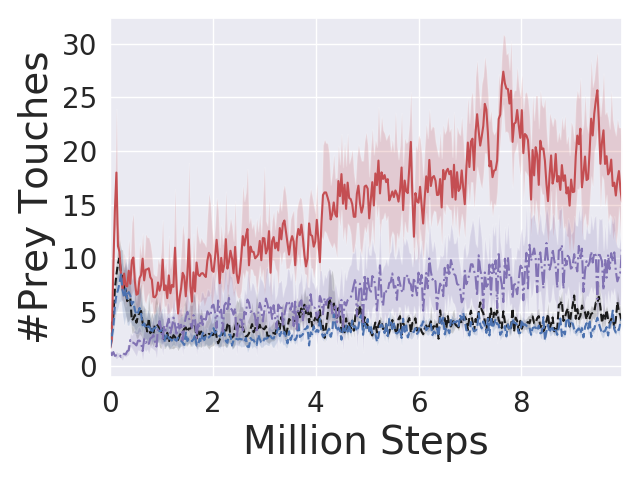}}
\subfigure[]{
\includegraphics[width=0.48\columnwidth]{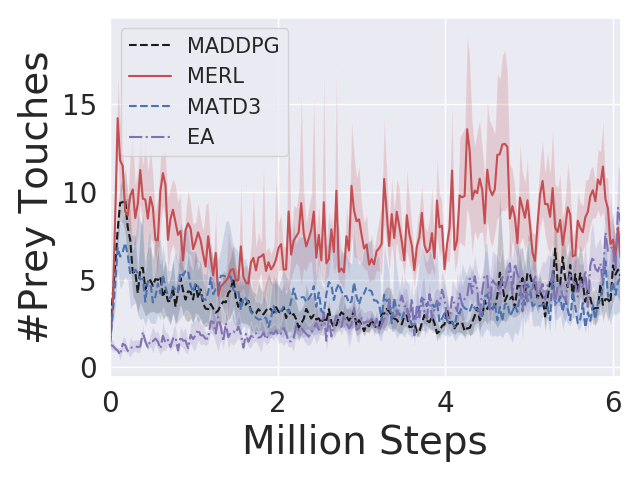}}
\caption{Results on Predator-Prey where the prey is (a) $30\%$ faster and (b) $100\%$ faster}
\label{fig:predator-prey}
\end{figure}

\textbf{Predator-Prey:} Figure \ref{fig:predator-prey} shows the comparative performance in controlling the team of predators in the Predator-Prey environment. Note that this is an adversarial environment where the prey dynamically adapts against the predators. The prey (considered as part of the environment in this analysis) uses DDPG to learn constantly against our team of predators. This is why predator performance (measured as number of prey touches) exhibits ebb and flow during learning. MERL outperforms MATD3, EA, and MADDPG across both simple and hard variations of the task. EA seems to be approaching MERL's performance, but is significantly slower to learn. This is an expected behavior for neuroevolutionary methods, which are known to be sample-inefficient. In contrast, MERL, by virtue of its fast policy-gradient components, learns significantly faster. 

\begin{figure}[h]
\subfigure[]{
\includegraphics[width=0.48\columnwidth]{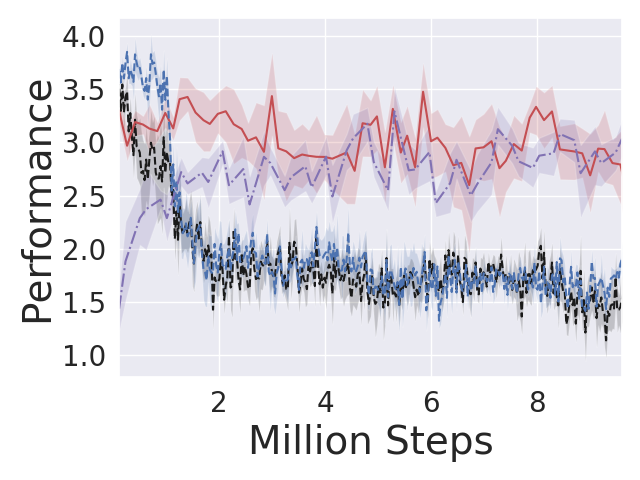}}
\subfigure[]{
\includegraphics[width=0.48\columnwidth]{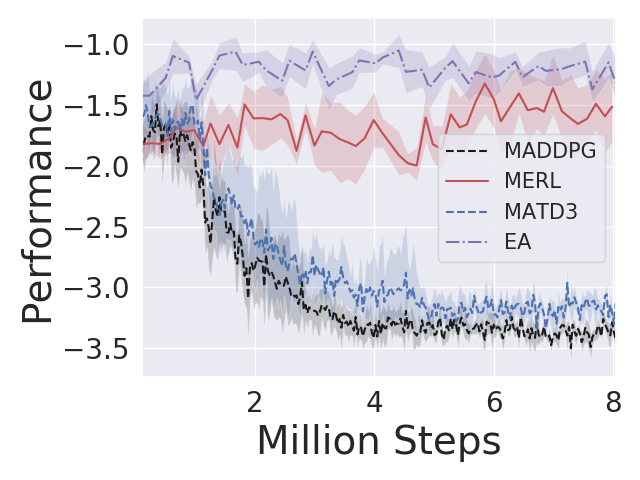}}
\caption{Results on (a) Physical Deception and (b) Keep-Away}
\label{fig:deception}
\vskip 0.1in
\end{figure}

\textbf{Physical Deception:} Figure \ref{fig:deception} (left) shows the comparative performance in controlling the team of agents in the Physical Deception environment. The performance here is largely based on how close the adversary comes to the target POI. Since the adversary starts out untrained, all compared algorithms start out with a fairly high score. As the adversary gradually learns to infer and move towards the target POI, MATD3 and MADDPG demonstrate a gradual decline in performance. However, MERL and EA are able to hold their performance by concocting effective counter-strategies in deceiving the adversary. EA reaches the same performance as MERL, but is slower to learn.

\textbf{Keep-Away:} Figure \ref{fig:deception} (right) show the comparative performance in Keep-Away. Similar to Physical Deception, MERL and EA are able to hold performance by attaining good counter-measures against the adversary while MATD3 and MADDPG fail to do so. However, EA slightly outperforms MERL on this task.

\begin{figure*}[t]
\subfigure[Coupling 1]{
\includegraphics[width=0.48\columnwidth]{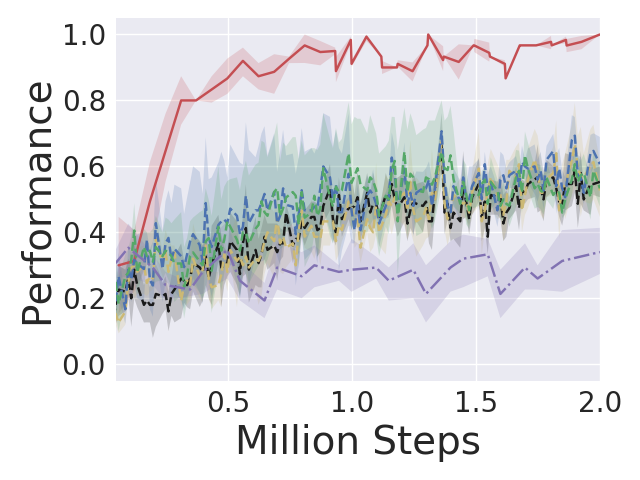}}
\subfigure[Coupling 3]{
\includegraphics[width=0.48\columnwidth]{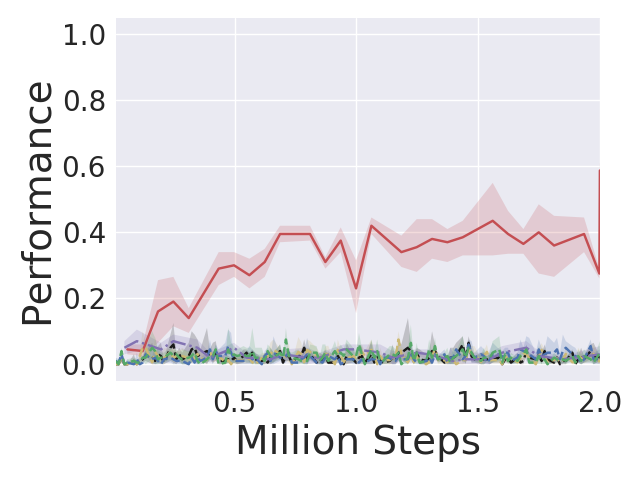}}
\subfigure[Coupling 7]{
\includegraphics[width=0.48\columnwidth]{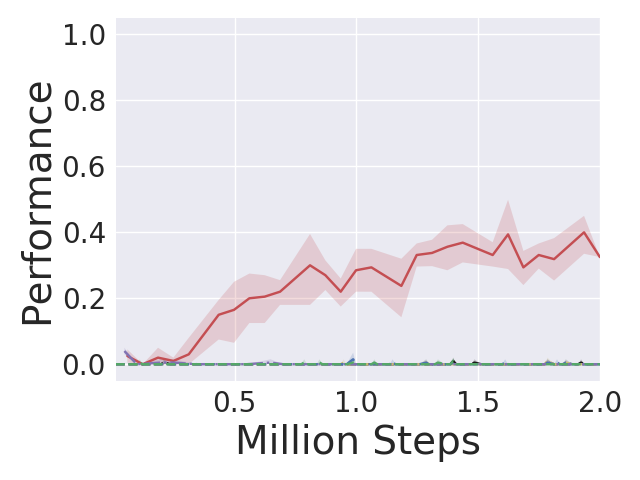}}
\subfigure[Legend]{
\includegraphics[width=0.48\columnwidth]{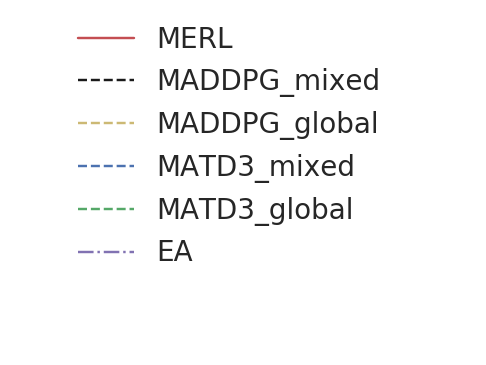}}
\caption{Performance on the Rover Domain.}
\label{fig:rover_domain_results}
\end{figure*}

\begin{figure}[t]
\centering
\includegraphics[width=0.85\columnwidth]{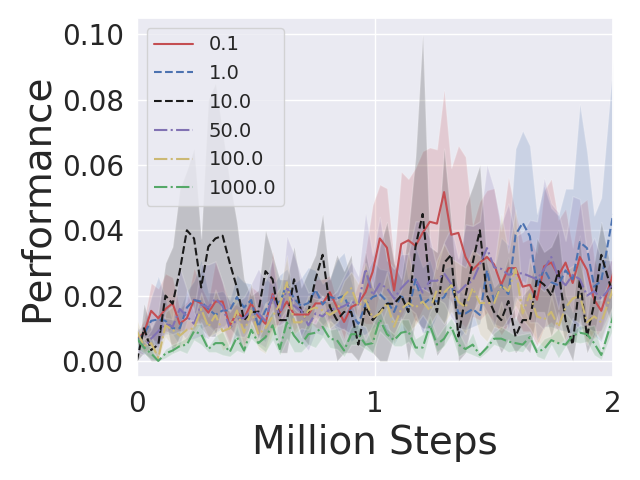}
\caption{MATD3's with varying scalarization coefficients}
\label{fig:scalarization}
\vskip -0.1in
\end{figure}

\textbf{Rover Domain:} Figure \ref{fig:rover_domain_results} shows the comparative performance of MERL, MADDPG, MATD3, and EA tested in the rover domain with coupling factors $1, 3$ and $7$. In order to benchmark against the proxy reward functions that use scalarized linear combinations, we test MADDPG and MATD3 with two variations of reward functions. \textit{Global} represents the scenario where the agents only receive the sparse team reward as their reinforcement signal. \textit{Mixed} represents the scenario where the agents receive a linear combination of the team-reward and agent-specific reward. Each reward is normalized before being combined. A weighing coefficient of $10$ is used to amplify the team-reward's influence in order to counter its sparsity. The weighing coefficient was tuned using a grid search (see Figure \ref{fig:scalarization}).

MERL significantly outperforms all baselines across all coupling requirements. The tested baselines clearly degrade quickly beyond a coupling of $3$. The increasing coupling requirement is equivalent to increasing difficulty in joint-space exploration and entanglement in the team objective. However, it does not increase the size of the state-space, complexity of perception, or navigation. This indicates that the degradation in performance is strictly due to the increase in complexity of the team objective. 

Notably, MERL is able to learn on coupling greater than $n=6$ where methods without explicit reward shaping have been shown to fail entirely \citep{rahmattalabi2016d++}. MERL successfully completes the task using the same set of information and coarse, unshaped reward functions as the other algorithms. This is mainly due to MERL's split-level approach, which allows it to leverage the agent-specific reward function to solve navigation and perception while concurrently using the team-reward function to learn team formation and effective coordination.

\textbf{Scalarization Coefficients for Mixed Rewards:} Figure \ref{fig:scalarization} shows the performance of MATD3 in optimizing mixed rewards computed with different coefficients used to amplify the team-reward relative to the agent-reward. The results demonstrate that finding a good balance between these two rewards through linear scalarization is difficult, as all values tested fail to make any progress in the task. This is because a static scalarization cannot capture the dynamic properties of \textit{which reward is important when} and instead leads to an ineffective proxy. In contrast, MERL is able to leverage both reward functions without the need to explicitly combine them either linearly or via more complex mixing functions.   

\begin{figure}[h]
\subfigure[MERL]{
\includegraphics[width=0.48\columnwidth]{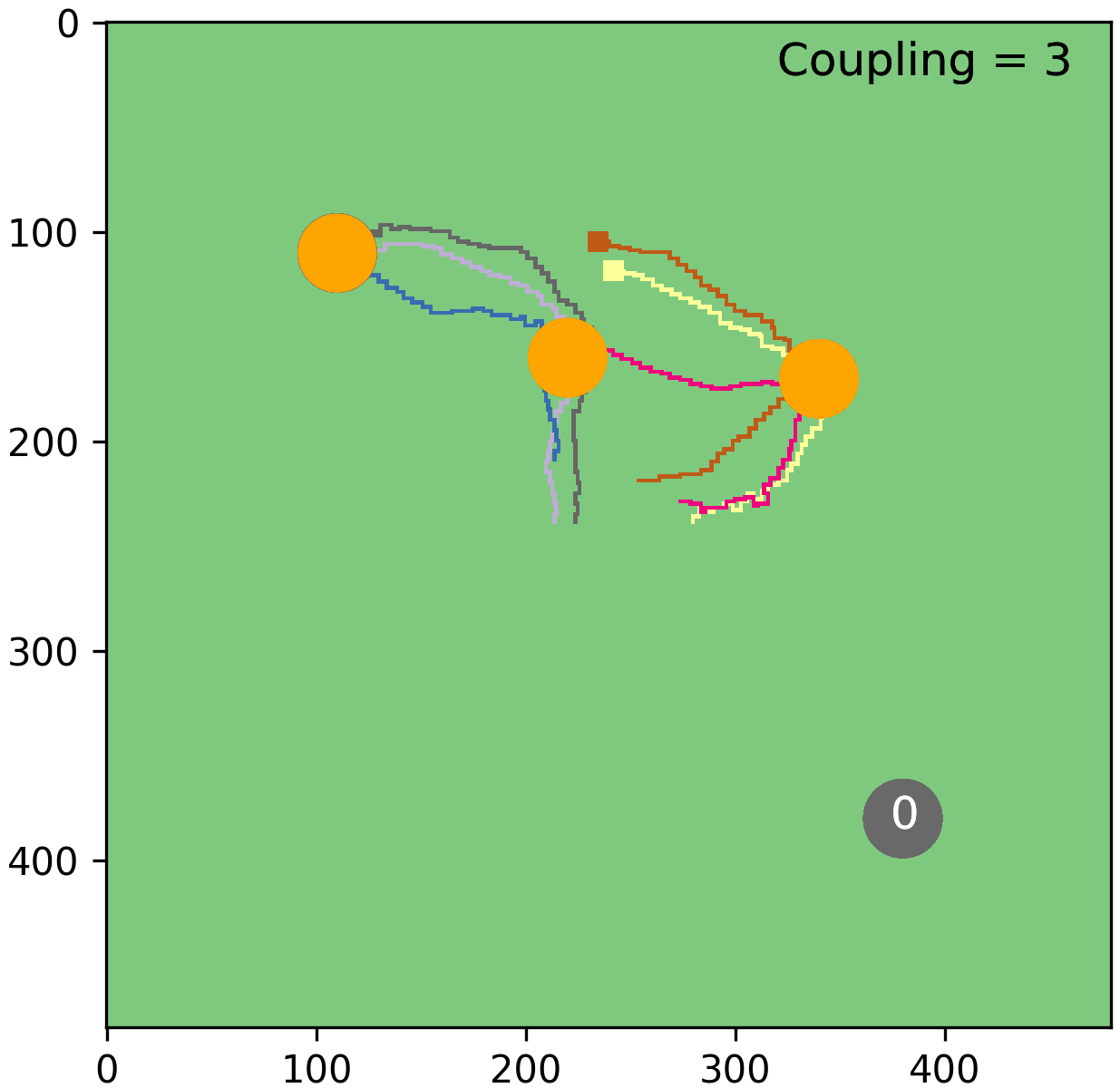}}
\subfigure[MADDPG]{
\includegraphics[width=0.48\columnwidth]{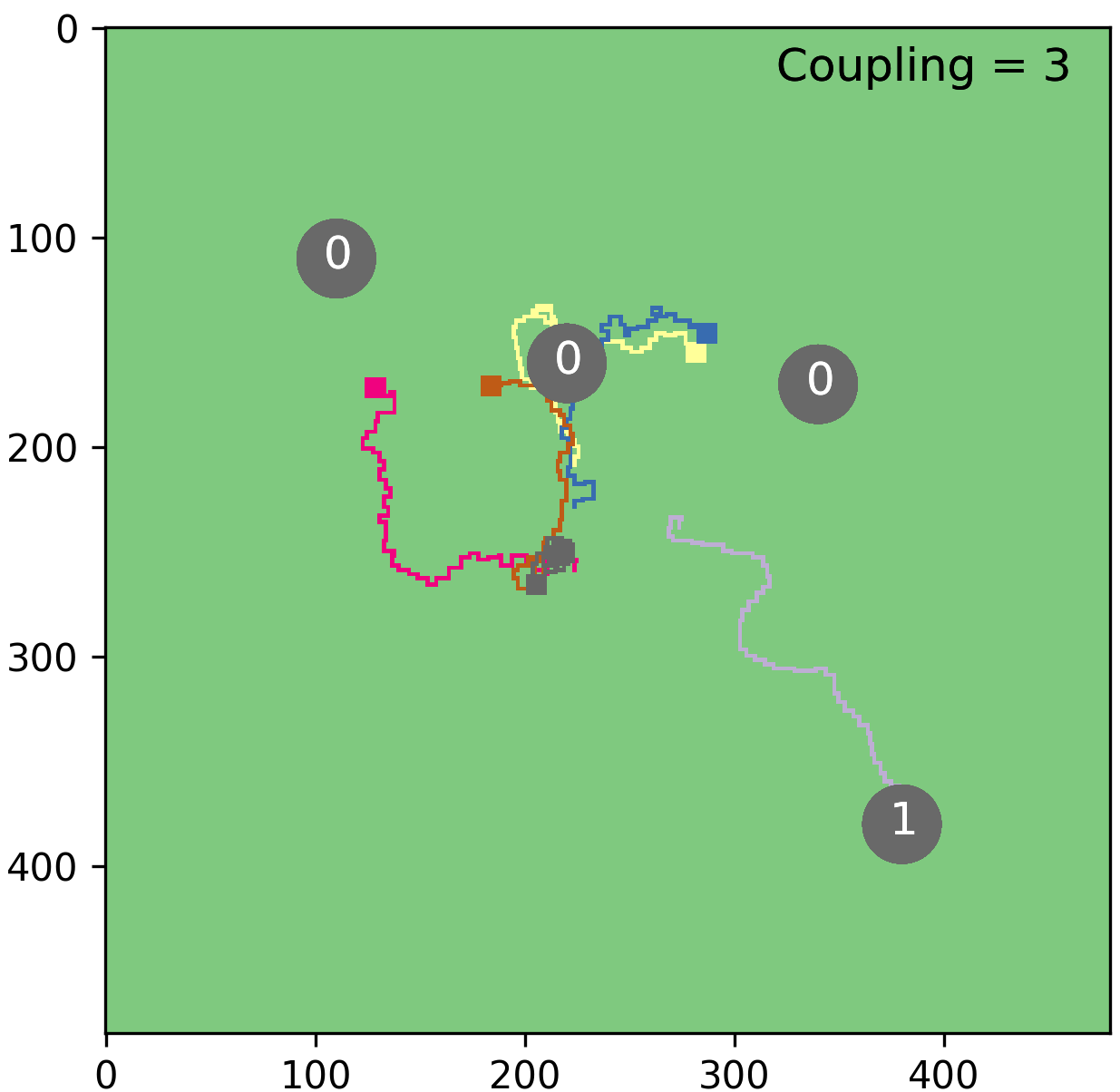}}
\caption{Agent trajectories for coupling = $3$. Golden/black circles are observed/unobserved POIs respectively }
\label{fig:rover_domain_viz}
\vskip 0.2in
\end{figure}

\textbf{Team Behaviors:} Figure \ref{fig:rover_domain_viz} illustrates the trajectories generated for the Rover Domain with a coupling of $n=3$ - the animations can also be viewed at our code repository online \footnote{\href{https://tinyurl.com/ugcycjk}{https://tinyurl.com/ugcycjk}}. The trajectories for a team fully trained with MERL is shown in Figure \ref{fig:rover_domain_viz} (a). Here, team formation and collaborative pursuit of the POIs is immediately apparent. Two teams of $3$ agents each form at the start of the episode. Further, the two teams also coordinate to pursue different POIs in order to maximize the team reward. While not perfect (the bottom POI is left unobserved), they do succeed in observing $3$ out of the $4$ POIs.  
\vspace{1em}

In contrast, MADDPG-mixed (shown in Figure \ref{fig:rover_domain_viz} (b)) fails to observe any POI. From the trajectories, it is apparent that the agents have successfully learned to perceive and navigate to reach POIs. However, they are unable to use this \emph{skill} towards fulfilling the team objective. Instead each agent is rather split on the objective that it is optimizing. Some agents seem to be in sole pursuit of POIs without any regard for team formation or collaboration while others seem to exhibit random movements. 
The primary reason for this is the mixed reward function that directly combines the agent-specific and team reward functions. Since the two reward functions have no guarantees of alignment across the state-space of the task, they invariably lead to learning these sub-optimal joint-behaviors that solve a certain form of scalarized mixed objective. In contrast, MERL by virtue of its bi-level optimization framework is able to leverage both reward functions without the need to explicitly combine them. This enables MERL to avoid these sub-optimal policies and solve the task without any reward shaping or manual tuning.

\begin{figure}[t]
\centering
\includegraphics[width=0.95\columnwidth]{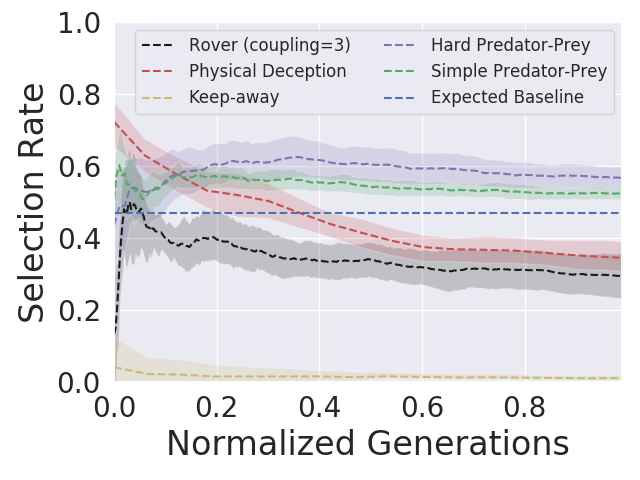}
\caption{Selection rate for migrating policies}
\label{fig:selection}
\vskip -0.2in
\end{figure}

\textbf{Conditional Selection Rate:} We ran experiments tracking whether the policies migrated from the policy gradient learners to the evolutionary population were selected or discarded during the subsequent selection process (Figure \ref{fig:selection}). Mathematically, this is equivalent to the conditional probability of selection ($sel$), given that the individual was a migrant this generation ($mig$), represented as $P (sel \mid mig)$. The expected selection rate $P (sel)$ under uniform random selection is $0.47$ (see Appendix B). Conditional selection rate above this baseline indicates that migration is a positive predictor for selection, and vice-versa. The conditional selection rate are distributed across both sides of this line, varying by task. The lowest selection rate is seen for Keep-away where evolution virtually discards all migrated individuals. This is consistent with the performance in Keep-Away (Figure \ref{fig:deception}b) where EA outperforms all other baselines while the policy-gradient methods struggle.

Both of the predator-prey tasks have consistently high conditional selection rate, indicating positive information transfer from the policy-gradient to the evolutionary population throughout training. This is consistent with Figure \ref{fig:predator-prey} where MERL outperforms all other baselines. For a dynamic environment such as predator-prey where the prey is consistently adapting against the agents, such information transfer is crucial for sustaining success. For physical deception, as well as the rover domain, the evolutionary population initially benefits heavily from the migrating individuals (consistent with Figure \ref{fig:deception}(a)). However, as these information propagates through the population, the marginal benefit from migration wanes through the course of learning. This form of adaptive information transfer is a key characteristic within MERL, enabling it to integrate two optimization processes towards maximizing the team objective without being misguided erroneously by either.   

\section{Conclusion}
 
 We introduced MERL, a hybrid multiagent reinforcement learning algorithm that leverages both agent-specific and team objectives by combining gradient-based and gradient-free optimization. MERL achieves this by using a fast policy-gradient optimizer to exploit dense agent-specific rewards while concurrently leveraging neuroevolution to tackle the team-objective. Periodic migration and a shared replay buffer ensure consistent information flow between the two processes. This enables MERL to leverage both modes of learning towards tackling the global team objective. 
 
Results demonstrate that MERL significantly outperforms MADDPG, the state-of-the-art MARL method, in a wide array of benchmarks. We also tested a modification of MADDPG to integrate TD3 - the state-of-the-art single-agent RL algorithm. These experiments demonstrate that the core improvements of MERL come from its ability to leverage team and agent-specific rewards without the need to explicitly combine them. This differentiates MERL from other approaches like reward scalarization and reward shaping that either require extensive manual tuning or can detrimentally change the MDP \citep{ng1999policy} itself. 

In this paper, MERL expended a constant amount of resources across its policy-gradient and EA components throughout training. Exploring automated methods to dynamically allocate computational resources among the two processes based on the conditional selection rate is an exciting thread for further investigation. Other future threads will explore MERL for settings such as Pommerman \citep{resnick2018pommerman}, StarCraft \citep{risi2017, vinyals2017starcraft}, RoboCup \cite{kitano1995robocup}, football \citep{kurach2019google} and general multi-reward settings such as multitask learning. 

\bibliographystyle{abbrvnat}
\bibliography{merl.bib}
\clearpage
\appendix

\section{Hyperparameters Description}

\begin{table}[h]
\caption{Hyperparameters used for Predator-Prey, Keep-away and Physical Deception}

\begin{tabular}{l|c|c} 
 \hline
 Hyperparameter & MERL & MATD3/MADDPG\\ 
 \hline \hline
    Population size                     & $10$ & N/A\\ 
    Rollout size                        & $10$                 & $10$\\ 
    Target weight                       & $0.01$  & $0.01$\\ 
    Actor Learning Rate                 & $0.01$ & $0.01$\\ 
    Critic Learning Rate                & $0.01$ & $0.01$\\ 
    Discount Factor                     & $0.95$ & $0.95$\\ 
    Replay Buffer Size                  &  $1e^{6}$ & $1e^{6}$\\ 
    Batch Size                          & $1024$ &$1024$\\ 
    Mutation Prob                       & $0.9$ & N/A\\ 
    Mutation Fraction                   &  $0.1$ & N/A\\ 
    Mutation Strength                   & $0.1$ & N/A\\          
    Super Mutation Prob                 & $0.05$& N/A \\ 
    Reset Mutation Prob                 & $0.05$& N/A \\ 
    Number of elites                    & $4$& N/A \\ 
    Exploration Policy                  & $\mathcal{N}(0, \sigma)$ & $\mathcal{N}(0, \sigma)$\\ 
    Exploration Noise                   & $0.4$ &$0.4$ \\ 
    Rollouts per fitness                & $10$ & N/A\\
    Actor Architecture                  & $[100, 100]$ & $[100, 100]$ \\
    Critic Architecture                 & $[100, 100]$ & $[300, 300]$ \\
    TD3 Noise Variance                  & $0.2$ & $0.2$ \\
    TD3 Noise Clip                      & $0.5$ & $0.5$ \\
    TD3 Update Freq                     & $2$ & $2$ \\

 \hline \hline
\end{tabular}
\label{merl_table_maddpg}
\vspace{2em}
\end{table}

Table \ref{merl_table_maddpg} details the hyperparameters used for MERL, MATD3, and MADDPG in tackling predator-prey and cooperative navigation. The hyperparmaeters were inherited from \cite{lowe2017multi} to match the original experiments for MADDPG and MATD3. The only exception to this was the use of hyperbolic tangent instead of Relu activation functions.

\begin{table}[h]
\caption{Hyperparameters used for Rover Domain}
\begin{tabular}{l|c|c} 
 \hline
 Hyperparameter & MERL & MATD3/MADDPG\\ 
 \hline \hline
    Population size                 & 10 & N/A\\ 
    Rollout size                    & $50$  & $50$\\ 
    Target weight                   & $1e^{-5}$  & $1e^{-5}$\\ 
    Actor Learning Rate             & $5e^{-5}$ & $5e^{-5}$\\ 
    Critic Learning Rate            & $1e^{-5}$ & $1e^{-5}$\\ 
    Discount Factor                 & $0.5$ & $0.97$\\ 
    Replay Buffer Size              &  $1e^{5}$ & $1e^{5}$\\ 
    Batch Size                      & $512$ &$512$\\ 
    Mutation Prob                   & $0.9$ & N/A\\ 
    Mutation Fraction               &  $0.1$ & N/A\\ 
    Mutation Strength               & $0.1$ & N/A\\                 
    Super Mutation Prob             & $0.05$& N/A \\ 
    Reset Mutation Prob             & $0.05$& N/A \\ 
    Number of elites                & $4$   & N/A \\ 
    Exploration Policy              & $\mathcal{N}(0, \sigma)$ & $\mathcal{N}(0, \sigma)$\\ 
    Exploration Noise $\sigma$      & $0.4$ &$0.4$ \\ 
    Rollouts per fitness $\xi$      & $10$ & N/A\\
    Actor Architecture              & $[100, 100]$ & $[100, 100]$ \\
    Critic Architecture             & $[100, 100]$ & $[300, 300]$ \\
    TD3 Noise variance              & $0.2$ & $0.2$ \\
    TD3 Noise Clip                  & $0.5$ & $0.5$ \\
    TD3 Update Frequency            & $2$ & $2$ \\

 \hline \hline
\end{tabular}
\label{merl_table}
\vspace{2em}
\end{table}

Table \ref{merl_table} details the hyperparameters used for MERL, MATD3, and MADDPG in the rover domain. The hyperparameters themselves are defined below:

\begin{itemize}

\item \textbf{Optimizer = Adam} \\
Adam optimizer was used to update both the actor and critic networks for all learners.

\item \textbf{Population size $M$} \\This parameter controls the number of different actors (policies) that are present in the evolutionary population. 

\item \textbf{Rollout size} \\This parameter controls the number of rollout workers (each running an episode of the task) per generation.

\textbf{Note:} The two parameters above (population size $k$ and rollout size) collectively modulates the proportion of exploration carried out through noise in the actor's \textit{parameter} space and its \textit{action} space.

\item \textbf{Target weight $\tau$} \\This parameter controls the magnitude of the soft update between the actors and critic networks, and their target counterparts.

\item \textbf{Actor Learning Rate} \\
This parameter controls the learning rate of the actor network.

\item \textbf{Critic Learning Rate} \\
This parameter controls the learning rate of the critic network.

\item \textbf{Discount Rate} \\
This parameter controls the discount rate used to compute the return optimized by policy gradient. 

\item \textbf{Replay Buffer Size} \\
This parameter controls the size of the replay buffer. After the buffer is filled, the oldest experiences are deleted in order to make room for new ones.

\item \textbf{Batch Size} \\
This parameters controls the batch size used to compute the gradients.

\item \textbf{Actor Activation Function } \\
Hyperbolic tangent was used as the activation function. 

\item \textbf{Critic Activation Function } \\
Hyperbolic tangent was used as the activation function. 

\item \textbf{Number of Elites} \\
This parameter controls the fraction of the population that are categorized as elites. Since an elite individual (actor) is shielded from the mutation step and preserved as it is, the elite fraction modulates the degree of exploration/exploitation within the evolutionary population. 

\item \textbf{Mutation Probability} \\
This parameter represents the probability that an actor goes through a mutation operation between generation.

\item \textbf{Mutation Fraction} \\
This parameter controls the fraction of the weights in a chosen actor (neural network) that are mutated, once the actor is chosen for mutation.

\item \textbf{Mutation Strength} \\
This parameter controls the standard deviation of the Gaussian operation that comprises mutation.

\item \textbf{Super Mutation Probability} \\
This parameter controls the probability that a super mutation (larger mutation) happens in place of a standard mutation.

\item \textbf{Reset Mutation Probability } \\
This parameter controls the probability a neural weight is instead reset between $\mathcal{N}(0,1)$ rather than being mutated.

\item \textbf{Exploration Noise} \\
This parameter controls the standard deviation of the Gaussian operation that comprise the noise added to the actor's actions during exploration by the learners (learner roll-outs). 

\item \textbf{TD3 Policy Noise Variance} \\
This parameter controls the standard deviation of the Gaussian operation that comprise the noise added to the policy output before applying the Bellman backup. This is often referred to as the magnitude of policy smoothing in TD3.

\item \textbf{TD3 Policy Noise Clip} \\
This parameter controls the maximum norm of the policy noise used to smooth the policy.

\item \textbf{TD3 Policy Update Frequency} \\
This parameter controls the number of critic updates per policy update in TD3.

\end{itemize}

\section{Expected Selection Rate}

We use a multi-step selection process inside MERL. First we select $e$ top-performing individuals as elites sequentially from the population without replacement. Then we conduct a tournament selection \cite{miller1995genetic} with tournament size $t$ with replacement from the entire population including the elites. $t$ is set to $3$ in this paper. The candidates selected from the tournament selection are pruned for duplicates and the resulting set is carried over as the offsprings for this generation. The combined set of offsprings and the elites represent the candidates selected for that generation of evolution.

The expected selection rate $P(s)$ is defined as the probability of selection for an individual if the selection was random. This is equivalent to conducting selection using a random ranking of the population where the fitness scores were ignored and a random number was assigned as an individual's fitness. Note that the selection rate is \textbf{not} the probability of selection for a random policy (individual with random neural network weights) inserted into the evolutionary population. The probability of selection for such a random policy would be extremely low as the other individuals in the population would be ranked significantly higher. 

In order to compute the expected selected rate, we need to compute it for the elite set and the offspring set. The expected selection rate for the elite set is given by $e/M$. The expected selection rate for the offspring set involves multiple rounds of tournament selection with replacement followed by pruning of duplicates to compute the combined set along with the elites. We computed this expectation empirically using an experiment with $1000000$ iterations and found the expected selection rate to be $0.47$.

\begin{figure*}[]
\centering
\subfigure[ES Population Sweep]{
\includegraphics[width=0.75\columnwidth]{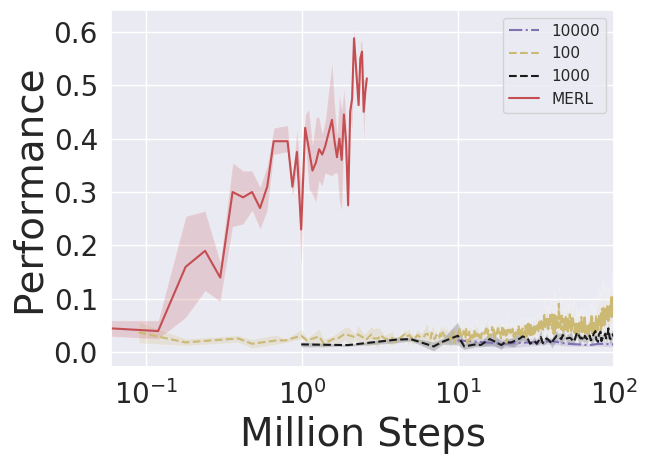}
}
\subfigure[ES Sigma Sweep]{
\includegraphics[width=0.75\columnwidth]{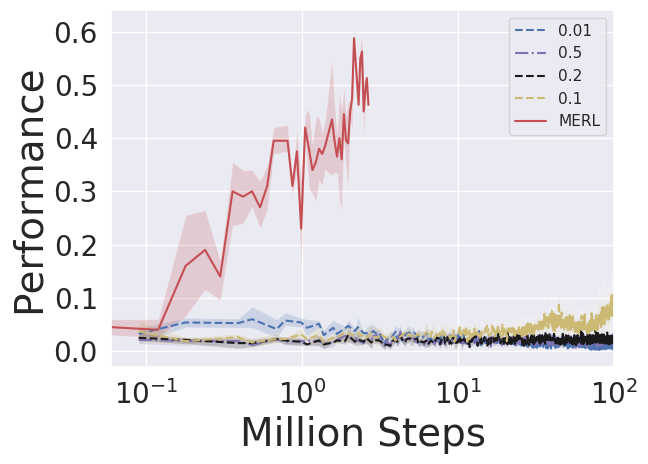}}

\caption{(a) Evolutionary Strategies population size sweep on the rover domain with a coupling of 3. (b)Evolutionary Strategies Noise magnitude (sigma) sweep on the rover domain with a coupling of 3}
\label{fig:es_pop}
\end{figure*}

\section{Extended EA Benchmarks}
\label{appendix:appendix_ea}

We conducted some additional experiments by varying the population sizes used for the EA baseline. The purpose of these experiments is to investigate if larger population sizes (as is the norm for EA algorithms) can alleviate the need for policy-gradient module within MERL.

Additionally, we also investigated Evolutionary Strategies (ES) \cite{salimans2017}, which has been widely adopted in the community in recent years. We perform hyperparameter sweeps to tune this baseline. All results are reported in the rover domain with a coupling of 3. 


\subsection{Evolutionary Strategies (ES)}

\paragraph{ES Population Sweep:}
Figure \ref{fig:es_pop}(left) compares ES with varying population sizes in the rover domain with a coupling of $3$. Sigma for all ES runs are set at $0.1$. Among the ES runs, a population size of $100$ yields the best results converging to $0.1$ in 100-millions frames. MERL (red) on the other hand is ran for 2-million frames and converges to $0.48$.

\paragraph{ES Noise Sweep:}
Apart from the population size, a key hyperparameter for ES is the variance of the perturbation factor (sigma). We run a parameter sweep for sigma and report results in Figure \ref{fig:es_pop}(right). We do not see a great deal of improvement with the change of sigma.

\vspace{1em}
\begin{figure}[H]
\centering
\includegraphics[width=0.75\columnwidth]{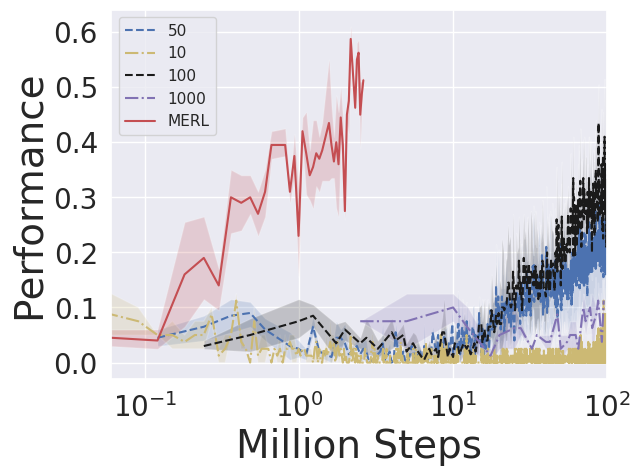}
\caption{Evolutionary Algorithm Population size sweep on the rover domain with a coupling of 3. MERL was run for 2-million steps while the other EA runs were ran for 100-million steps.}
\label{fig:ea_pop}
\end{figure}

\subsection{EA Population Size}
Next, we conduct an experiment to evaluate the efficacy of different population sizes from $10$-$1,000$ for the Evolutionary algorithm used in the paper. All results are reported for the rover domain with a coupling factor of 3 and are illustrated in Figure \ref{fig:ea_pop}. The best EA performance was found for a population size of 100 reaching ~$0.3$ in $100$-million time steps. Compare this to MERL which reaches a performance of ~$0.48$ in only $2$ million time steps. This demonstrates a key thesis behid MERL - the efficacy of the guided evolution approach over purely evolutionary approaches. 

\section{Rollout Methodology}

Algorithm \ref{alg:merl_episode} describes an episode of rollout under MERL detailing the connections between the local reward, global reward, and the associated replay buffer.

\begin{algorithm}[]
    \caption{Rollout}
    \label{alg:merl_episode} 
\begin{algorithmic}[1]
\FUNCTION{Rollout($\pi$, $\mathcal{R}$, noise, $\xi$)}
    \STATE $fitness = 0$
    \FOR{j = 1:$\xi$}
    	\STATE Reset environment and get initial joint state $js$
        \WHILE{env is not done}
            \STATE Initialize an empty list of joint action $ja$ = []
            \FOR{Each agent (actor head) $\pi^k$ $\in$ $\pi$ and $s_k$ $in$ $js$} 
                \STATE $ja$ $\Leftarrow$ $ja$ $\cup$ $\pi^k(s_k|\theta^{\pi^k}) + noise_t$ 
            \ENDFOR
            
            \STATE Execute $ja$ and observe joint local reward $jl$, global reward $g$ and joint next state $js'$ 
            
            \FOR{Each Replay Buffer $\mathcal{R}_k$ $\in$ $\mathcal{R}$ and $s_k$, $a_k$, $l_k$, $s'_k$ in $js$, $ja$, $jl$, $js'$}
                \STATE Append transition $(s_k, a_k, l_k, s'_k)$ to $R_k$ 
            \ENDFOR
            \STATE $js = js'$
            \IF{env is done:} 
                \STATE $fitness \leftarrow g$
            \ENDIF
        \ENDWHILE
    \ENDFOR
    \STATE Return $\frac{fitness}{\xi}$, $\mathcal{R}$
\ENDFUNCTION

\end{algorithmic}
\end{algorithm}

\end{document}